\documentclass[10pt,twocolumn,letterpaper]{article}
\usepackage{times,xspace}
\usepackage[dvipsnames]{xcolor}
\usepackage{graphicx,amsmath,amssymb,booktabs}
\usepackage[numbers,sort&compress]{natbib}
\usepackage[format=plain,labelformat=simple,labelsep=period,font=small,compatibility=false]{caption}
\usepackage[hyphens]{url}
\usepackage[shortlabels,inline]{enumitem}
\setlist[itemize,enumerate]{noitemsep,leftmargin=*,topsep=0em}
\usepackage{cuted,placeins}
\usepackage[pagebackref,breaklinks,colorlinks,hypertexnames=false]{hyperref}
\definecolor{linkblue}{rgb}{0.21,0.49,0.74}
\hypersetup{allcolors=linkblue,
  pdftitle={SERA: Towards Reliable Medical Image Segmentation through Agentic Error Review},
  pdfauthor={Delin An, Mingqi Xu, Pengfei Gu, Jian-Xun Wang}}

\newcommand{\etal}{\emph{et al}.\null\xspace}
\newcommand{\papertitle}{SERA: Towards Reliable Medical Image Segmentation\\ through Agentic Error Review}
\title{\papertitle}
\author{Delin An\textsuperscript{1}, Mingqi Xu\textsuperscript{1},
  Pengfei Gu\textsuperscript{2}, and Jian-Xun Wang\textsuperscript{1}}
\date{}

\makeatletter
\renewcommand{\@maketitle}{%
  \newpage\null\vskip .375in
  \begin{center}
    {\Large\bfseries\@title\par}
    \vspace*{24pt}{\large\lineskip .5em
      \begin{tabular}[t]{c}
        \smash{\begin{tabular}{@{}c@{}}
           \@author\\
           {\normalsize\textsuperscript{1}Cornell University\qquad
           \textsuperscript{2}The University of Texas Rio Grande Valley}\\
           {\small\ttfamily\{da568, mx263, jw2837\}@cornell.edu\qquad pengfei.gu01@utrgv.edu}
        \end{tabular}}
      \end{tabular}\par}
    \vskip .5em\vspace*{12pt}
  \end{center}}
\renewenvironment{abstract}{\centerline{\large\bfseries Abstract}%
  \vspace*{12pt}\noindent\itshape\ignorespaces}{\vspace*{12pt}}
\renewcommand{\@seccntformat}[1]{\csname the#1\endcsname.\nobreakspace}
\renewcommand{\section}{\@startsection{section}{1}{\z@}%
  {-10pt plus -2pt minus -2pt}{7pt}{\large\bfseries}}
\font\subheadingfont=ptmb scaled 1100
\renewcommand{\subsection}{\@startsection{subsection}{2}{\z@}%
  {-8pt plus -2pt minus -2pt}{5pt}{\subheadingfont}}
\makeatother

\begin{document}
\maketitle
\begin{strip}
\centering
\includegraphics[width=\textwidth]{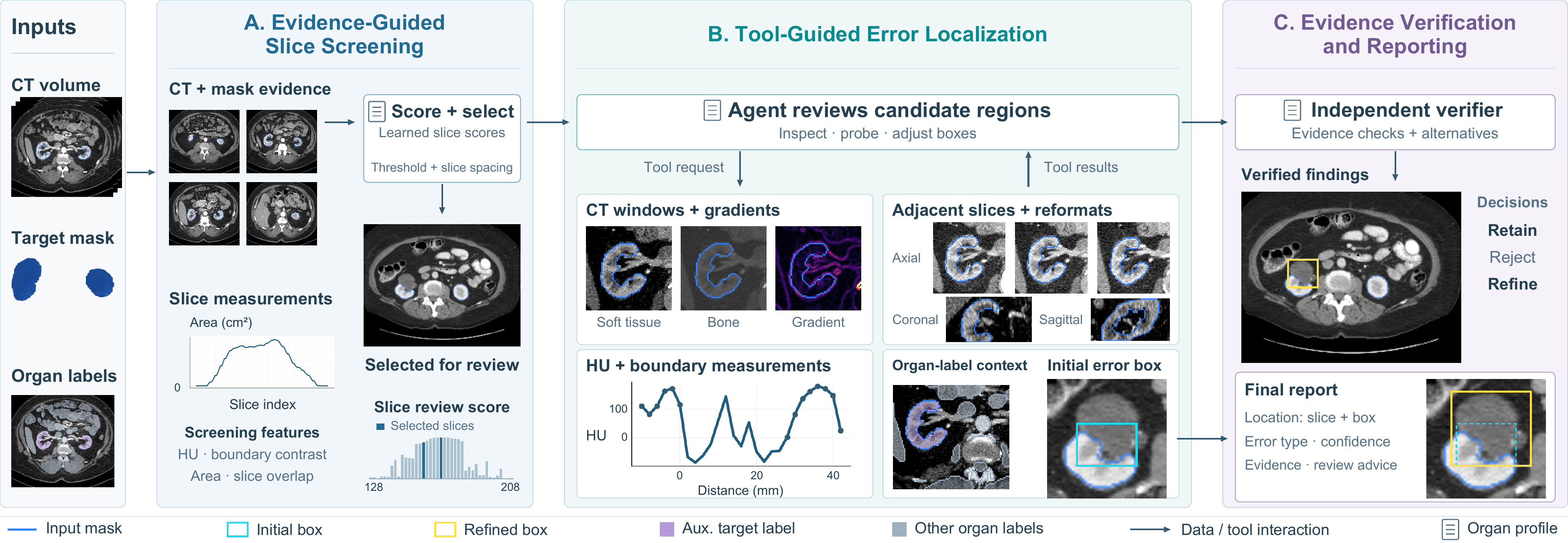}
\captionsetup{hypcap=false}
\captionof{figure}{
\textbf{Overview of SERA.}
(a) Evidence-guided slice screening combines local error evidence and cross-slice consistency with agent-guided selection.
(b) Tool-guided error localization uses image inspection and regional measurements to refine bounding boxes around suspected errors.
(c) Evidence verification combines numerical checks with independent review to produce a structured report.
}
\label{fig:pipeline}

\end{strip}
\begin{abstract}
Deep learning has been widely adopted for organ segmentation in medical images, yet its predictions can still contain errors, such as missing target tissue or incorrectly including surrounding structures. Experts typically identify these errors by examining tissue contrast, anatomical continuity, and organ-specific characteristics across multiple views. However, manual review is time-consuming, and existing automated methods often require experts to interpret the detected errors. 
We propose SERA, an agentic framework for detecting and localizing errors in 3D organ segmentations. SERA follows a structured review process that combines rapid screening, organ-specific analysis, and evidence-based verification. First, a lightweight logistic regression model scores slices using features inspired by expert review, including boundary contrast, mask area changes, and inter-slice overlap; the agent sets a threshold to select slices for inspection. For each selected slice, SERA generates initial bounding boxes based on organ-specific intensity and geometric features. The agent then examines windowed images, multiplanar views, and tissue measurements, using organ profiles to distinguish segmentation errors from normal anatomical structures. It further refines the bounding boxes through additional observations and geometric tools to improve localization accuracy. Finally, a verification step resolves ambiguous cases and reduces incorrect predictions.
SERA outputs the location, type, confidence, and supporting evidence for each detected error. We evaluate SERA across multiple organs and CT and MRI datasets, demonstrating its effectiveness in detecting and localizing segmentation errors.
\end{abstract}

\section{Introduction}
\label{sec:intro}

Deep learning has advanced a wide range of medical imaging tasks, including segmentation~\cite{Chen2025}, classification~\cite{Nie2026}, and image restoration~\cite{Cao2026}.
In segmentation, both manually drawn and automatically generated contours can omit target tissue, extend into neighboring structures, or contain disconnected fragments~\cite{Brooks2024,Kart2022}.
To detect these errors, experts compare the contour with the source image in axial, coronal, and sagittal views, checking whether it follows the organ boundary and covers the anatomical extent~\cite{RCR2024,Kart2022}.
Neighboring contours also guide manual delineation by showing how the target continues across slices~\cite{Zhang2022}.

Local image contrast and knowledge of the target help reviewers interpret suspicious regions.
In CT, reviewers use tissue-appropriate windows to assess interfaces and anatomical landmarks to distinguish adjacent structures~\cite{RCR2024,Chupetlovska2025,Jabbour2014}.
They also apply organ-specific contouring conventions. For example, the Radiation Therapy Oncology Group's abdominal CT guidelines call for including portal vein branches within the liver contour~\cite{Jabbour2014}.
Tissue that looks different from the surrounding parenchyma can therefore be included in the mask under the applicable protocol.
Reviewing each structure across slices and planes requires dedicated expert time, even when few regions need correction~\cite{RCR2024,Gibbons-JMRS23}.

Statistical quality checks can flag atypical contours using shape descriptors or organ-specific CT intensity statistics~\cite{Wooten2023,Brooks2024}.
Deep networks can estimate slice-level Dice or produce voxelwise error maps~\cite{Fournel-MIA21,Qiu-MIA26,Specktor-Fadida-MIA25}.
Supervised detectors learn from discrepancies with reference masks, so which discrepancies count as errors depends on the reference contours and the tolerance for boundary variation~\cite{Qiu-MIA26,Specktor-Fadida-MIA25}.
These maps can highlight a suspected omission, but the reviewer still has to determine whether the region is continuous with the target, belongs to a neighboring organ, or is excluded by the contouring protocol.
Tool-using multimodal agents can make these checks explicit and choose which observation to acquire next~\cite{Yao-ICLR23,Fallahpour-ICML25}.
For example, a weak boundary can prompt a different CT window, while uncertain tissue identity can prompt a coronal view to trace the region's connection to the target.

We introduce \textbf{SERA}, an agentic framework for reviewing predicted organ segmentations in volumetric medical images (Fig.~\ref{fig:pipeline}).
SERA translates features from expert review into measurable features. Changes in mask area and overlap summarize continuity across slices, while disconnected components, tissue intensities, and boundary gradients capture local shape and image features.
Organ-specific profiles supply intensity ranges and geometric scales calibrated on reference cases, together with the annotation conventions used to interpret candidate errors.
SERA uses these features to rank slices, while the agent sets the screening threshold and performs detailed inspection of selected slices.
This work makes three main contributions.

\begin{enumerate}
    \item \textbf{Feature-based logistic regression for slice screening.} We decouple error detection into slice-level screening and subsequent fine-grained localization, reducing the complexity of direct voxel-level error prediction. A lightweight logistic regression model ranks slices using regional error features, mask geometry, and cross-slice consistency. The agent sets a case-specific threshold to select slices for review. We further suppress slices that repeat the same suspected error, focusing review on distinct regions.
    \item \textbf{Tool- and profile-guided box localization.} We use organ-specific profiles to construct local error evidence and fit initial boxes to the resulting regions. Windowed views, multiplanar inspection, and tissue measurements help localize suspected omissions and excess tissue.
    \item \textbf{Agentic box verification and refinement.} The agent checks competing tissue interpretations with follow-up measurements and uses geometric tools to adjust boxes to the error evidence or connected tissue. Numerical checks and a separate verifier reassess ambiguous findings and unsupported rejections. Each reported box includes an error type and supporting observations.
\end{enumerate}

\section{Related Work}
\label{sec:related_work}

\noindent\textbf{Conventional quality control.}
Statistical contour checks flag geometric outliers using previously annotated cases.
Hui \etal~\cite{Hui2018} compare volumetric descriptors of organs at risk with their distributions in a reference cohort.
Reverse classification accuracy~\cite{Valindria2017} uses the candidate mask to construct a reverse classifier, then takes its best agreement on annotated reference images as a proxy for mask quality.
Both methods flag suspect organs for review.

\noindent\textbf{Deep learning for error detection.}
Uncertainty methods derive a quality signal from the segmenter's predictions.
Zenk \etal~\cite{Zenk-MIA25} found that pairwise Dice agreement across ensemble predictions helped identify poor masks under distribution shifts.
Voxelwise uncertainty provides more spatial detail, though its calibration can vary across patients~\cite{Jungo-MICCAI19}.
Rhee \etal~\cite{Rhee2019} trained an independent convolutional segmenter for head and neck organs and flagged candidate contours with low overlap against its predictions.
An error shared by both segmenters can still yield a high overlap score.
Quality regressors estimate overlap directly from the image and candidate mask.
Fournel \etal~\cite{Fournel-MIA21} use a convolutional network to estimate quality at the slice level and combine slice measurements to estimate volume-level overlap.
The choice of regression target determines which aspect of segmentation quality the model estimates.

Spatial detectors estimate where errors occur within a slice or volume.
Zaman \etal~\cite{Zaman2023} learn image features near a segmented boundary to identify erroneous surface patches.
QCResUNet~\cite{Qiu-MIA26} couples a 3D regression encoder with a decoder that predicts tissue-specific voxelwise error maps.
The authors observed that localization accuracy could drop on high-quality masks, where the remaining errors can be small.
SegQC~\cite{Specktor-Fadida-MIA25} derives multiple quality metrics from voxelwise error probabilities and extracts local regions while filtering narrow contour discrepancies.
These detectors learn from differences between predicted and reference masks, so the definition of an error depends on the reference and the tolerance used for contour variation.
Reconstruction provides another route to quality assessment.
For example, nnQC~\cite{Marciano2026} conditions latent diffusion on image features and relative slice position to generate a comparison mask.
The same reconstruction supports several quality metrics, whose accuracy depends on how well it represents the patient's anatomy.

\noindent\textbf{Visual agents.}
For natural images, V*~\cite{Wu2024} uses target and contextual features to prioritize image patches and returns crops for detailed inspection.
Visual Sketchpad~\cite{Hu2024} feeds the model views annotated with boxes or masks to support spatial reasoning.
DeepEyes~\cite{Zheng2026} learns when and where to crop an image through reinforcement learning.
In medical imaging, MedRAX~\cite{Fallahpour-ICML25} selects classification, localization, and segmentation tools to answer questions about chest radiographs.
For volumetric CT, 3DMedAgent~\cite{Wang2026} combines predicted organ masks with lesion features to select slices for radiology questions.
MedSAM-Agent~\cite{Liu2026} uses successive point and box prompts to refine masks on 2D medical images.
RaMA~\cite{Zhang2027} uses a multimodal model to check cardiac MRI pseudo-labels and prompt three SAM variants, whose consensus masks support target-domain self-training.
These evaluations measure answer accuracy or the quality of masks the agent produces.
Reviewing an existing volumetric mask requires localizing missing or excess tissue and linking each finding to observations that a reviewer can verify against the original scan.

\section{Method}
\label{sec:method}

Given a CT or MRI volume $X\in\mathbb{R}^{H\times W\times Z}$, a predicted target mask $S\in\{0,1\}^{H\times W\times Z}$, and a target organ category $o$, SERA identifies suspected segmentation errors and explains where and why they require review.
An auxiliary organ label map $L$ provides anatomical context.
Because MRI lacks a standardized intensity scale~\cite{Nyul2000}, we center body intensities at their median and scale their interquartile range to $100$.
All inputs share a common image grid, with voxel spacing $\boldsymbol{\Delta}=(\Delta_x,\Delta_y,\Delta_z)$.
We use $z$ for an axial slice, $j$ for a candidate region within that slice, and $t$ for an agent turn.
Each finding records the slice index, box coordinates, error direction, confidence, and supporting evidence.

The review proceeds from volume-level screening to local inspection and evidence verification.
First, an evidence-based screening module concentrates the agent's attention on informative slices while limiting repeated inspection of the same structure.
Second, the agent uses image and measurement tools to examine candidate regions and select a spatial extent for each suspected error.
Third, numerical checks and a separate verification pass assess both reported findings and rejected candidates before producing the final report.
This organization combines lightweight slice scoring with agent-guided selection and detailed inspection.
An organ-specific profile $\Theta_o$, estimated offline from reference patients, supplies intensity ranges, geometric scales, annotation conventions, and scoring parameters shared across these steps.
This design is inspired by expert assessment of tissue appearance and organ anatomy under the relevant contouring protocol~\cite{Chupetlovska2025,Jabbour2014}.

\subsection{Evidence-Guided Slice Screening}
\label{sec:slice_screening}

A volume contains many similar slices, whereas a segmentation error may occupy only a limited anatomical region.
Inspecting every slice with an LLM would require substantial effort to review repeated information.
We therefore rank slices by local error evidence and cross-slice consistency, then select a compact set for detailed review.
Screening samples axial slices at approximately $3$\, mm intervals over the longitudinal extent of the target mask and any auxiliary target prediction, with $10$\, mm of context at each end.

\noindent\textbf{Constructing error evidence.}
For each sampled slice, we extract spatial evidence of potential errors from the image, mask, neighboring masks, and organ labels.
Potential omissions include enclosed holes, tissue continuing beyond a mask boundary without a corresponding image edge, and tissue covered on neighboring slices but absent from the current mask.
Potential excess includes atypical intensities within the mask, disconnected fragments, narrow protrusions, and mask boundaries that extend beyond image edges.
Disagreement with auxiliary predictions provides additional candidates for inspection.
The profile $\Theta_o$ adapts intensity and geometric criteria to the target and represents annotation conventions, such as structures routinely included within an organ outline.
These observations indicate where the mask warrants inspection; the subsequent image review resolves their interpretation.

Spatially connected regions with compatible error evidence form candidates $R_{zj}$, each with a proposed direction $d_{zj}\in\{\mathrm{under},\mathrm{over}\}$ and an initial box.
Enclosed holes are kept separate from external omissions to preserve distinct spatial hypotheses.
A feature vector $\mathbf{f}_{zj}$ describes the region's area, types of supporting evidence, intensity difference from the mask interior and reference tissue, boundary contrast, contact with the mask, and auxiliary label composition.
The geometric and intensity features are inspired by expert checks of anatomical boundaries, tissue contrast, and disconnected fragments~\cite{Kart2022,Chupetlovska2025}.
Its estimated error probability is
\begin{equation}
    q_{zj}=g_{\theta_o}(\mathbf{f}_{zj}),
    \label{eq:candidate_probability}
\end{equation}
where $g_{\theta_o}$ is an organ-specific logistic or gradient-boosted classifier selected using patient-grouped cross-validation on reference cases.
The parameters $\theta_o$ are part of $\Theta_o$.

\noindent\textbf{Scoring slices.}
We aggregate candidate evidence with the mask's cross-slice behavior into a slice descriptor $\boldsymbol{\phi}_z$.
It includes mask area, area changes and overlap with neighboring masks, relative position within the target extent, candidate counts and probabilities, and disagreement with auxiliary target labels.
Area changes and neighboring-mask overlap serve as computational proxies for the cross-slice continuity features used in manual contouring and review~\cite{Zhang2022,RCR2024}.
A logistic model estimates the priority of reviewing the slice:
\begin{equation}
    p_z=\sigma\!\left(\mathbf{w}_o^{\mathsf T}
    \widetilde{\boldsymbol{\phi}}_z+\beta_o\right),
    \label{eq:slice_probability}
\end{equation}
where $\mathbf{w}_o$ and $\beta_o$ are learned scoring parameters, $\sigma$ is the sigmoid function, and $\widetilde{\boldsymbol{\phi}}_z$ is standardized using reference-set statistics.
The candidate and slice models share a spatial supervision criterion: on reference cases, a positive candidate box overlaps at least $50$\,mm$^2$ of an error component of at least $150$\,mm$^2$, with at least $10\%$ of its area occupied by errors of the proposed direction.
A slice is positive when it contains at least one such candidate.
Thus, $p_z$ prioritizes slices with an inspectable error candidate.

\noindent\textbf{Selecting complementary slices.}
Slices are considered in descending order of $p_z$. The agent dynamically determines an organ-specific threshold $\tau_o$ based on the characteristics of the current case to select potentially erroneous slices. To avoid spending the review budget on repeated cross-sections of one error, we also suppress a slice when its leading candidate has the same direction as an already selected candidate, lies within $40$ mm along the axial axis, and overlaps at least half of the smaller box in-plane. The resulting set $\mathcal{Z}_{\mathrm{sel}}$ guides the agent to distinct candidate regions.

\subsection{Tool-Guided Error Localization}
\label{sec:error_localization}

Slice selection determines where to allocate attention, but precise localization requires resolving the extent and identity of suspicious tissue.
For each $z\in\mathcal{Z}_{\mathrm{sel}}$, the agent receives a volume overview with the mask and numbered candidate boxes, a table of regional measurements and probabilities, and the target's annotation guidance.
Region identifiers link visual observations, tool requests, and reported findings to the same spatial locations.

\noindent\textbf{Initializing spatial proposals.}
Each initial box $b_{zj}^{(0)}$ is fitted to its evidence region.
Available extents include the full region, its portion near the mask boundary, and its compact core; a policy estimated from reference cases selects the initial extent based on the target, error direction, and dominant evidence type.
The box includes a physical margin and a minimum viewing size, so small candidate regions remain inspectable without losing their surrounding anatomy.
Boxes use global slice coordinates $(x_0,y_0,x_1,y_1)$ with exclusive upper bounds.
Displayed crops preserve physical aspect ratio and show global coordinate ticks, maintaining a consistent correspondence between enlarged views and the original slice.

\noindent\textbf{Inspecting with tools.}
The agent chooses which candidate to inspect and which observation is needed next.
The \texttt{zoom} tool provides enlarged views and gradient maps, using CT windows or MRI display ranges from the target profile.
The \texttt{views} tool provides neighboring axial slices and coronal and sagittal reformats through the candidate, letting the agent assess anatomical continuity beyond a single plane.
The \texttt{hu\_profile} tool measures intensities and boundary gradients and samples a line from the mask interior toward the candidate.
The \texttt{identity} tool compares label composition, tissue intensities, interfaces, and three-dimensional continuity with the target and nearby structures.
These tools are implemented as Python functions to help distinguish tissue missed by the mask from neighboring organs and to distinguish true extensions of the target from leakage across tissue boundaries during segmentation.

Let $\mathcal{H}_z^{(t)}$ denote the accumulated observations and tool results for slice $z$.
The LLM policy $\pi_\psi$, with backbone parameters $\psi$, selects an action $a_z^{(t)}$ that is either a tool request or a final decision.
For a tool action, the interaction is
\begin{equation}
\begin{aligned}
    a_z^{(t)} &= \pi_\psi(\mathcal{H}_z^{(t)};\Theta_o),\\
    \mathcal{H}_z^{(t+1)} &= \mathcal{H}_z^{(t)}\oplus
    \bigl(a_z^{(t)},\mathcal{T}_z(a_z^{(t)})\bigr),
\end{aligned}
\label{eq:agent_interaction}
\end{equation}
where $\mathcal{T}_z$ executes the requested tools against the volume and current candidate state, and $\oplus$ appends the interaction to the history.
This feedback lets later requests depend on earlier evidence: an ambiguous boundary can trigger a different window, while uncertain tissue identity can trigger a cross-plane inspection.
The agent can also open a new candidate with \texttt{probe} when the overview reveals a suspicious structure outside the initial proposals.

\noindent\textbf{Refining the error extent.}
The agent uses \texttt{adjust\_box} to select among fitting the evidence region, fitting connected tissue, or tightening the current extent.
Fitting the evidence preserves a localized defect; fitting tissue is useful when the visible omission extends beyond the initial evidence region.
For tissue fitting, the evidence pixels seed constrained region growing.
Let $\overline{X}_z$ be the smoothed slice, $G_z$ its gradient magnitude in physical coordinates, and $m_{zj}$ the median intensity of the seeds.
Growth is restricted to
\begin{equation}
\begin{aligned}
    \Omega_{zj}=\bigl\{\mathbf{u}:\;&
    |\overline{X}_z(\mathbf{u})-m_{zj}|\leq\epsilon_{zj},\\
    &G_z(\mathbf{u})<\eta_z\bigr\}\cap D_{zj},
\end{aligned}
\label{eq:tissue_growth}
\end{equation}
where $\mathbf{u}$ is an image location. For CT, $\epsilon_{zj}$ is the larger of $25$ HU and $2.5$ times the seed intensity's median absolute deviation, and $\eta_z$ is the larger of $15$ HU/mm and the slice's $85$th gradient percentile.
For MRI, the same numerical rules use normalized MR units and MR units/mm, respectively.
The domain $D_{zj}$ lies outside the mask for an omission and inside it for an excess, excluding tissue assigned to other organs by the auxiliary map.
The original seeds are preserved, growth has a bounded physical radius, and only components connected to the seeds are retained.

If $U_{zj}^{(t)}$ is the extent chosen by the agent's geometric operation, the updated box is
\begin{equation}
    b_{zj}^{(t+1)}=\operatorname{Box}_{\boldsymbol{\Delta}}\!
    \left(U_{zj}^{(t)}\right),
    \label{eq:box_update}
\end{equation}
where $\operatorname{Box}_{\boldsymbol{\Delta}}$ computes the axis-aligned bounds with the physical margin and minimum size.
The tool returns the updated view and measurements for further inspection.
The agent selects the tissue extent, and the tool reports its coordinates on the original image grid.

\subsection{Evidence Verification and Reporting}
\label{sec:evidence_verification}

A plausible interpretation may still confuse neighboring tissues, and rejecting a suspicious region too early may hide a genuine omission.
SERA therefore checks both positive findings and rejection decisions before reporting them.
Each proposed finding includes an error direction, confidence level, image evidence, measurement evidence, the strongest alternative explanation, and a review instruction.
A rejection identifies its basis, such as another organ, a separating interface, or partial-volume effects.
The reference prior must explicitly address candidates recommended by it as findings or rejections, making otherwise overlooked candidates available for verification.

\noindent\textbf{Checking claims against measurements.}
Numerical checks compare the agent's explanation with the measured tissue properties.
For example, an excess claim based on auxiliary label disagreement is checked for intensity contrast, an internal edge, or, in CT, fat within the region.
An explanation that attributes tissue to another organ is checked against that organ's labels, contact, and intensity relationship.
Rejection checks also use the reference annotation convention, preventing an included structure from being dismissed solely because it differs from textbook organ parenchyma.
These checks link verbal judgments to the evidence the tools actually return.

\noindent\textbf{Verifying competing interpretations.}
A separate LLM call re-examines non-obvious findings, high-confidence findings with weak reference support, and unsupported rejections.
The verifier receives regional images, measurements, and reference guidance, without the reviewing agent's earlier conclusion or rationale.
It answers explicit questions about labels, intensity, the interface with the target, the interface with neighboring structures, and three-dimensional continuity, then compares alternative tissue interpretations.
Its error score $v_{zj}\in[0,1]$ is combined with the reference probability:
\begin{equation}
    \widehat{q}_{zj}=(1-\lambda)q_{zj}+\lambda v_{zj},
    \qquad \lambda=0.5.
    \label{eq:verification_score}
\end{equation}
For a newly probed region without a reference probability, the decision uses $v_{zj}$ directly.
The combined score controls retention and restoration: the default thresholds are $0.30$ to retain an existing finding and $0.50$ to restore a rejected candidate.
For a rejected candidate with strong reference support ($q_{zj}\geq0.60$), a verification score $v_{zj}\geq0.25$ also permits restoration.
The verifier can also select the evidence extent or the connected tissue extent, linking its interpretation to the reported location.

\noindent\textbf{Producing a concise report.}
After verification, overlapping findings are deduplicated.
Findings are classified as \emph{obvious}, \emph{probable}, or \emph{requiring review}.
To limit review burden, we retain at most one unresolved suggestion when no higher-confidence finding remains.
Each final record contains the slice index, global box coordinates, error direction, evidence, alternative explanation, and review instruction.
Measurements are recomputed for the final box, and the tool history and decision source are saved alongside it.
The resulting report directs expert attention to specific tissue regions with traceable reasons for inspection.

\noindent\textbf{LLM instantiation.}
The reviewing agent and verifier share a configurable vision-language backbone.
Local review allows up to four agent turns, with up to four tool calls per turn and a final decision required on the last turn.
For comparisons between LLMs, we replace the backbone in both roles while fixing the cases, selected slices, reference profiles, prompts, tool interfaces, interaction budgets, and post-processing rules.
We save findings before and after verification to track how each backbone changes the report.

\section{Experiments}
\label{sec:experiments}

\subsection{Experimental Setup and Evaluation Protocol}
\label{sec:experimental_setup}

\noindent\textbf{Datasets and input masks.}
We evaluate SERA on nine datasets, including six CT and three MRI datasets. The CT datasets comprise AortaSeg24~\cite{Imran2026}, KiTS19~\cite{Heller2021}, MSD Liver and MSD Spleen~\cite{Antonelli2022}, Pancreas-CT~\cite{Roth2016}, and TotalSegmentator~\cite{Wasserthal-RAI23}. The MRI datasets include AMOS pancreas~\cite{Ji2022}, CHAOS~\cite{Kavur2021}, and MSD Heart~\cite{Antonelli2022}. All input masks are automatically generated predictions, not ground-truth annotations. Specifically, we use pretrained TotalSegmentator~\cite{Wasserthal-RAI23} for all CT datasets, TotalSegmentator MRI for AMOS and CHAOS, and SCALA~\cite{Rege2026} for left atrium segmentation in MSD Heart. SERA takes these predicted masks and their corresponding medical images as inputs to identify and localize segmentation errors. Ground-truth annotations from test patients are used only for evaluation.

\noindent\textbf{Patient-level split.}
We assign approximately 30\% of the prepared patients in each dataset to testing. Patients used for tool development are excluded from test selection, and all targets and sequences from one patient share a partition. The remaining patients supply training and validation data. Each method is evaluated on the same scan--target pairs, including all axial slices of each volume.

\noindent\textbf{Baselines.}
We compare SERA with nnU-Net~\cite{Isensee-NM21}, QCResUNet~\cite{Qiu-MIA26}, and RaMA~\cite{Zhang2027}. We train nnU-Net as an error detection baseline, taking the medical image and its predicted segmentation mask as input to produce a voxel-level error map. The supervised baselines produce dense error maps, which we convert to axial error boxes. RaMA uses refined pseudo-labels for target-domain self-training. For fixed-mask review, we run its refinement stage with frozen models. GPT-4o mini selects slices for refinement and prompts SAM 3, MedSAM 2, and SAM-Med2D. Each refiner reverts to the input mask if its Dice with that mask falls below $0.50$. Voxels added by the majority vote indicate missing tissue, and removed voxels indicate excess tissue. SERA uses GPT-6 Astra unless stated otherwise.

\noindent\textbf{Box extraction.}
We threshold baseline error probabilities at $0.50$ on the evaluation grid. Error voxels outside the input mask indicate missing tissue, and those inside indicate excess tissue. For each type, we fit a tight axis-aligned box to each 4-connected component of at least $25\,\mathrm{mm}^2$. We apply the same component rule to errors derived from the input and reference masks. We use SERA's predicted boxes and error types directly.

\noindent\textbf{Metrics and aggregation.}
Precision ($P$) and recall ($R$) measure error-slice detection. A true positive is a reference-positive slice flagged by the method; a false positive is a flagged slice with no reference error; and a false negative is a reference-positive slice that is missed. Thus, $P=\mathrm{TP}/(\mathrm{TP}+\mathrm{FP})$ and $R=\mathrm{TP}/(\mathrm{TP}+\mathrm{FN})$. For each dataset, we compute $F_1=2PR/(P+R)$ from its mean precision and recall, with SD obtained by first-order propagation of their variability and covariance.

Box mIoU measures the mean overlap between predicted and reference error boxes. Error-type accuracy measures agreement between predicted and reference categories, namely missing or excess tissue. All quality scores are reported in percent as mean $\pm$ SD. Macro summaries weight the nine dataset--target groups equally and report the sample SD across their means.

\subsection{Baseline Comparison}
\label{sec:baseline_comparison}

\begin{table*}[tp]
    \centering
    \footnotesize
    \setlength{\tabcolsep}{1.5pt}
    \caption{Error-slice detection and box localization on nine CT and MRI dataset. Scores are mean $\pm$ SD (\%); $P$, $R$, and $F_1$ concern error-slice detection, mIoU concerns error boxes, and Type denotes error-type accuracy. Bold marks the highest mean among the four methods for each metric and dataset.}
    \label{tab:volume_localization}
    \begin{tabular*}{\textwidth}{@{\extracolsep{\fill}}l*{10}{r}@{}}
        \toprule
        \textit{(a)} & \multicolumn{5}{c}{nnU-Net} & \multicolumn{5}{c}{QCResUNet} \\
        \cmidrule(lr){2-6}\cmidrule(l){7-11}
        Dataset & $P\uparrow$ & $R\uparrow$ & $F_1\uparrow$ & mIoU $\uparrow$ & Type $\uparrow$ & $P\uparrow$ & $R\uparrow$ & $F_1\uparrow$ & mIoU $\uparrow$ & Type $\uparrow$ \\
        \midrule
        AortaSeg24 (CT) & $60.0\!\pm\!9.9$ & $53.8\!\pm\!11.6$ & $56.7\!\pm\!9.1$ & $27.0\!\pm\!10.5$ & $55.1\!\pm\!10.3$ & $59.6\!\pm\!10.3$ & $57.9\!\pm\!11.3$ & $58.7\!\pm\!9.0$ & $30.1\!\pm\!10.2$ & $49.6\!\pm\!10.3$ \\
        KiTS19 & $72.7\!\pm\!7.3$ & $69.1\!\pm\!8.2$ & $70.9\!\pm\!6.6$ & $41.3\!\pm\!7.3$ & $65.4\!\pm\!7.3$ & $71.3\!\pm\!7.3$ & $73.0\!\pm\!7.8$ & $72.1\!\pm\!6.4$ & $45.1\!\pm\!7.2$ & $60.6\!\pm\!7.6$ \\
        MSD Liver & $78.2\!\pm\!5.7$ & $75.6\!\pm\!6.4$ & $76.9\!\pm\!5.1$ & $53.3\!\pm\!5.8$ & $72.4\!\pm\!6.0$ & $75.9\!\pm\!5.6$ & $78.9\!\pm\!6.2$ & $77.4\!\pm\!4.9$ & $57.4\!\pm\!5.9$ & $67.9\!\pm\!6.4$ \\
        MSD Spleen & $80.3\!\pm\!4.8$ & $77.9\!\pm\!5.5$ & $79.1\!\pm\!4.3$ & $54.5\!\pm\!5.2$ & $73.8\!\pm\!5.5$ & $78.2\!\pm\!4.8$ & $81.5\!\pm\!5.6$ & $79.8\!\pm\!4.3$ & $59.0\!\pm\!5.4$ & $69.6\!\pm\!5.7$ \\
        Pancreas-CT & $63.7\!\pm\!8.2$ & $57.1\!\pm\!9.7$ & $60.2\!\pm\!7.6$ & $33.0\!\pm\!9.3$ & $60.9\!\pm\!9.4$ & $63.1\!\pm\!8.6$ & $61.3\!\pm\!9.9$ & $62.2\!\pm\!7.7$ & $36.5\!\pm\!9.3$ & $55.9\!\pm\!9.4$ \\
        TotalSegmentator & $77.3\!\pm\!6.0$ & $72.1\!\pm\!7.1$ & $74.6\!\pm\!5.6$ & $53.8\!\pm\!6.5$ & $72.8\!\pm\!6.4$ & $75.0\!\pm\!6.3$ & $75.4\!\pm\!7.0$ & $75.2\!\pm\!5.7$ & $57.8\!\pm\!6.3$ & $68.2\!\pm\!6.4$ \\
        \midrule
        AMOS (pancreas) (MRI) & $57.9\!\pm\!10.6$ & $49.8\!\pm\!12.0$ & $53.5\!\pm\!9.4$ & $25.7\!\pm\!10.7$ & $54.1\!\pm\!10.5$ & $57.6\!\pm\!10.6$ & $54.1\!\pm\!11.4$ & $55.8\!\pm\!9.0$ & $28.7\!\pm\!10.4$ & $48.6\!\pm\!10.9$ \\
        CHAOS & $74.9\!\pm\!6.7$ & $69.0\!\pm\!7.4$ & $71.8\!\pm\!6.0$ & $50.1\!\pm\!6.7$ & $69.8\!\pm\!6.9$ & $73.0\!\pm\!6.5$ & $72.6\!\pm\!7.2$ & $72.8\!\pm\!5.7$ & $54.2\!\pm\!6.8$ & $65.1\!\pm\!7.3$ \\
        MSD Heart & $72.2\!\pm\!7.1$ & $67.8\!\pm\!8.0$ & $69.9\!\pm\!6.4$ & $48.3\!\pm\!7.5$ & $69.7\!\pm\!7.9$ & $70.9\!\pm\!7.1$ & $71.7\!\pm\!8.1$ & $71.3\!\pm\!6.4$ & $52.4\!\pm\!7.8$ & $65.3\!\pm\!8.3$ \\
        \midrule
        Macro average & $70.8\!\pm\!8.2$ & $65.8\!\pm\!9.9$ & $68.2\!\pm\!9.1$ & $43.0\!\pm\!11.7$ & $66.0\!\pm\!7.6$ & $69.4\!\pm\!7.5$ & $69.6\!\pm\!9.6$ & $69.5\!\pm\!8.5$ & $46.8\!\pm\!12.2$ & $61.2\!\pm\!8.0$ \\
        \midrule
        \textit{(b)} & \multicolumn{5}{c}{RaMA} & \multicolumn{5}{c}{SERA} \\
        \cmidrule(lr){2-6}\cmidrule(l){7-11}
        Dataset & $P\uparrow$ & $R\uparrow$ & $F_1\uparrow$ & mIoU $\uparrow$ & Type $\uparrow$ & $P\uparrow$ & $R\uparrow$ & $F_1\uparrow$ & mIoU $\uparrow$ & Type $\uparrow$ \\
        \midrule
        AortaSeg24 (CT) & $61.9\!\pm\!9.8$ & $20.9\!\pm\!7.6$ & $31.2\!\pm\!8.9$ & $34.2\!\pm\!11.3$ & $42.9\!\pm\!11.4$ & $\mathbf{64.5}\!\pm\!7.3$ & $\mathbf{62.7}\!\pm\!8.1$ & $\mathbf{63.6}\!\pm\!6.6$ & $\mathbf{52.3}\!\pm\!7.9$ & $\mathbf{71.3}\!\pm\!7.0$ \\
        KiTS19 & $73.9\!\pm\!6.8$ & $36.3\!\pm\!5.3$ & $48.7\!\pm\!5.4$ & $47.4\!\pm\!8.3$ & $52.2\!\pm\!8.6$ & $\mathbf{78.1}\!\pm\!5.1$ & $\mathbf{77.5}\!\pm\!6.0$ & $\mathbf{77.8}\!\pm\!4.7$ & $\mathbf{66.7}\!\pm\!6.0$ & $\mathbf{81.4}\!\pm\!5.2$ \\
        MSD Liver & $79.6\!\pm\!5.4$ & $43.3\!\pm\!4.4$ & $56.1\!\pm\!4.4$ & $59.0\!\pm\!7.0$ & $58.9\!\pm\!7.2$ & $\mathbf{83.6}\!\pm\!4.3$ & $\mathbf{83.4}\!\pm\!5.0$ & $\mathbf{83.5}\!\pm\!4.0$ & $\mathbf{78.7}\!\pm\!5.0$ & $\mathbf{88.3}\!\pm\!4.2$ \\
        MSD Spleen & $82.2\!\pm\!4.8$ & $46.0\!\pm\!4.0$ & $59.0\!\pm\!3.9$ & $60.5\!\pm\!6.2$ & $60.5\!\pm\!6.1$ & $\mathbf{86.0}\!\pm\!3.9$ & $\mathbf{85.4}\!\pm\!4.5$ & $\mathbf{85.7}\!\pm\!3.7$ & $\mathbf{79.8}\!\pm\!4.2$ & $\mathbf{89.6}\!\pm\!3.5$ \\
        Pancreas-CT & $65.3\!\pm\!8.6$ & $24.1\!\pm\!6.9$ & $35.2\!\pm\!7.8$ & $39.8\!\pm\!10.3$ & $48.2\!\pm\!9.9$ & $\mathbf{68.8}\!\pm\!6.7$ & $\mathbf{66.2}\!\pm\!7.3$ & $\mathbf{67.5}\!\pm\!5.9$ & $\mathbf{58.6}\!\pm\!6.8$ & $\mathbf{77.1}\!\pm\!5.8$ \\
        TotalSegmentator & $78.6\!\pm\!6.1$ & $39.6\!\pm\!4.7$ & $52.7\!\pm\!4.9$ & $59.4\!\pm\!7.0$ & $59.2\!\pm\!7.0$ & $\mathbf{82.9}\!\pm\!4.5$ & $\mathbf{80.3}\!\pm\!5.0$ & $\mathbf{81.6}\!\pm\!4.1$ & $\mathbf{79.2}\!\pm\!4.8$ & $\mathbf{88.7}\!\pm\!4.3$ \\
        \midrule
        AMOS (pancreas) (MRI) & $59.6\!\pm\!9.9$ & $16.5\!\pm\!7.7$ & $25.8\!\pm\!9.7$ & $32.8\!\pm\!11.9$ & $41.9\!\pm\!12.4$ & $\mathbf{62.7}\!\pm\!7.5$ & $\mathbf{59.3}\!\pm\!8.5$ & $\mathbf{61.0}\!\pm\!6.7$ & $\mathbf{51.3}\!\pm\!8.6$ & $\mathbf{70.5}\!\pm\!7.6$ \\
        CHAOS & $76.6\!\pm\!6.2$ & $36.7\!\pm\!5.1$ & $49.6\!\pm\!5.2$ & $56.2\!\pm\!8.0$ & $56.6\!\pm\!8.2$ & $\mathbf{79.8}\!\pm\!4.9$ & $\mathbf{76.8}\!\pm\!5.8$ & $\mathbf{78.3}\!\pm\!4.6$ & $\mathbf{75.2}\!\pm\!5.7$ & $\mathbf{85.4}\!\pm\!4.8$ \\
        MSD Heart & $73.9\!\pm\!7.0$ & $35.4\!\pm\!5.8$ & $47.9\!\pm\!5.9$ & $54.4\!\pm\!9.1$ & $56.6\!\pm\!8.9$ & $\mathbf{77.6}\!\pm\!5.6$ & $\mathbf{76.2}\!\pm\!6.5$ & $\mathbf{76.9}\!\pm\!5.2$ & $\mathbf{73.8}\!\pm\!6.2$ & $\mathbf{85.7}\!\pm\!5.1$ \\
        \midrule
        Macro average & $72.4\!\pm\!8.2$ & $33.2\!\pm\!10.3$ & $45.1\!\pm\!11.6$ & $49.3\!\pm\!11.1$ & $53.0\!\pm\!7.1$ & $\mathbf{76.0}\!\pm\!8.6$ & $\mathbf{74.2}\!\pm\!9.3$ & $\mathbf{75.1}\!\pm\!8.9$ & $\mathbf{68.4}\!\pm\!11.6$ & $\mathbf{82.0}\!\pm\!7.4$ \\
        \bottomrule
    \end{tabular*}
\end{table*}

\noindent\textbf{Detection and localization.}
Table~\ref{tab:volume_localization} compares error-slice detection, box localization, and error-type prediction. SERA achieves the highest mean in all five metrics on every group. Its macro precision, recall, and $F_1$ are 76.0\%, 74.2\%, and 75.1\%, respectively. The macro $F_1$ exceeds nnU-Net (68.2\%), QCResUNet (69.5\%), and RaMA (45.1\%) by 6.9, 5.6, and 30.0 percentage points. RaMA attains 72.4\% precision but only 33.2\% recall, indicating that it misses a larger proportion of reference-positive slices than the other methods.

SERA reaches 68.4\% box mIoU, exceeding the strongest localization baseline, RaMA (49.3\%), by 19.1 percentage points. Its error-type accuracy is 82.0\%, compared with 66.0\% for the strongest baseline, nnU-Net, a gain of 16.0 points. Thus, SERA's advantage extends from detecting erroneous slices to localizing errors and distinguishing missing from excess tissue.

\noindent\textbf{Variation across datasets.}
SERA's $F_1$ gains over QCResUNet, the strongest detection baseline on every group, range from 4.9 to 6.4 percentage points. Absolute performance nevertheless varies across datasets: SERA's $F_1$ ranges from 61.0\% on AMOS pancreas to 85.7\% on MSD Spleen. Box mIoU is also lower on AMOS pancreas (51.3\%) and AortaSeg24 (52.3\%) than on MSD Spleen (79.8\%) and TotalSegmentator (79.2\%). The improvements are therefore consistent across the evaluated groups, while substantial differences in detection and localization performance remain.

\begin{figure*}[tp]
    \centering
    \includegraphics[width=0.85\textwidth]{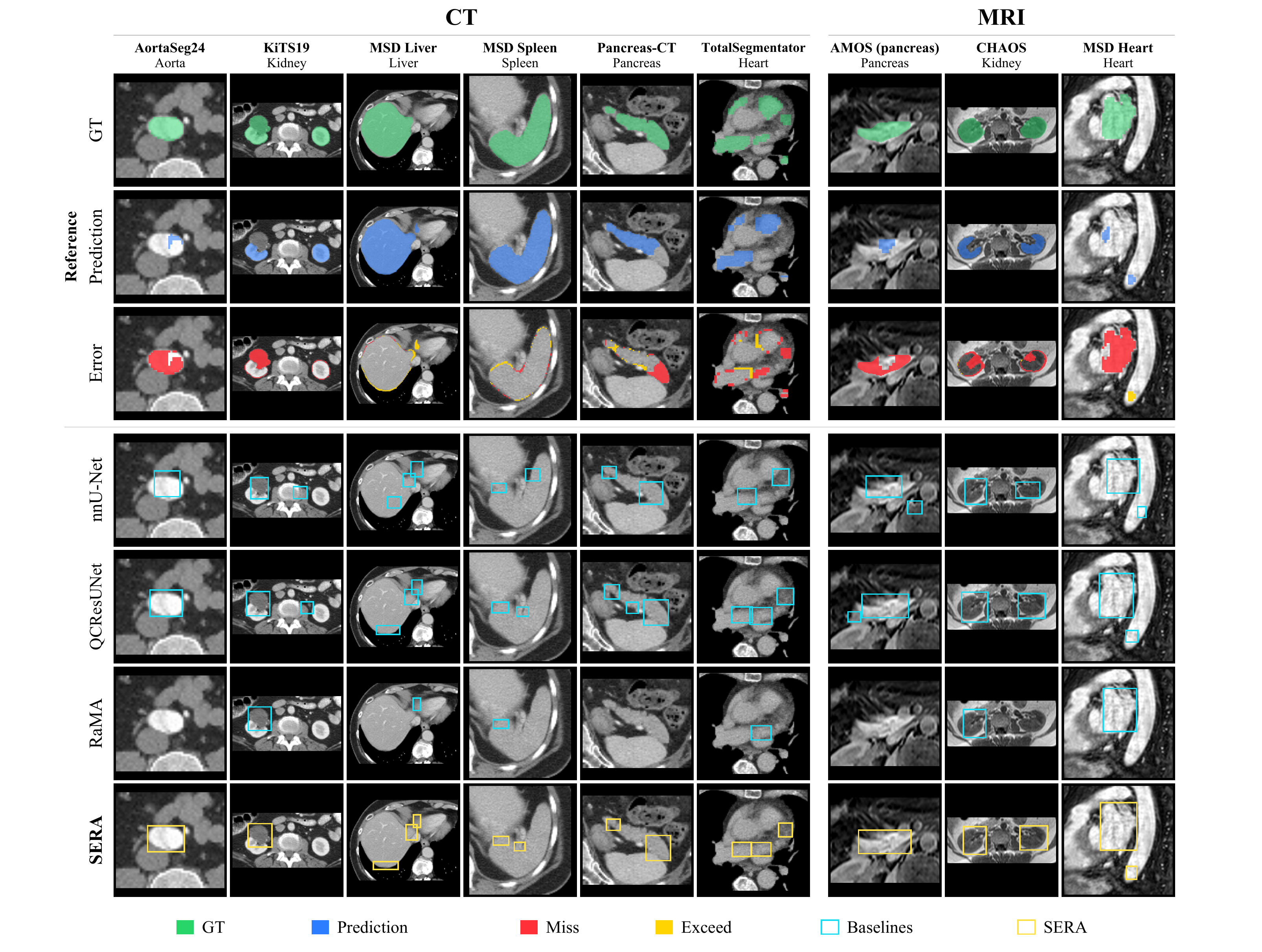}
    \caption{Qualitative comparison of CT and MRI slices, with the same image and crop within each column. The top three rows show reference masks (green), input segmentations (blue), and missing (red) or excess (yellow) tissue.}
    \label{fig:qualitative}
\end{figure*}

\noindent\textbf{Qualitative comparison.}
Figure~\ref{fig:qualitative} presents qualitative comparisons on matched CT and MRI slices. As shown in the figure, SERA produces bounding boxes that align more closely with the actual error regions, with more accurate locations and sizes than the baselines. These qualitative observations are consistent with the quantitative results.

\subsection{LLM Agent Comparison and Inference Cost}
\label{sec:backbone_comparison}

\begin{table*}[tp]
    \centering
    \footnotesize
    \setlength{\tabcolsep}{1.5pt}
    \caption{LLM agent comparison within SERA. Quality scores are macro mean $\pm$ SD across the nine datasets (\%), with the highest mean in each metric in bold. Tokens and API costs are averaged per complete agent-reviewed slice. Input denotes uncached input, Cached denotes cache-read tokens, and Total sums input, cached, and output tokens. Relative cost is normalized to GPT-6 Astra.}
    \label{tab:backbones}
    \begin{tabular*}{\textwidth}{@{\extracolsep{\fill}}l*{11}{r}@{}}
        \toprule
        & \multicolumn{5}{c}{Detection and localization} & \multicolumn{4}{c}{Tokens per reviewed slice} & \multicolumn{2}{c}{API cost} \\
        \cmidrule(lr){2-6}\cmidrule(lr){7-10}\cmidrule(l){11-12}
        LLM agent & $P\uparrow$ & $R\uparrow$ & $F_1\uparrow$ & mIoU $\uparrow$ & Type $\uparrow$ & Input & Cached & Output & Total & USD & Relative \\
        \midrule
        GPT-6 Astra & $76.0\!\pm\!8.6$ & $\mathbf{74.2}\!\pm\!9.3$ & $\mathbf{75.1}\!\pm\!8.9$ & $\mathbf{68.4}\!\pm\!11.6$ & $\mathbf{82.0}\!\pm\!7.4$ & 10,500 & 8,400 & 6,850 & 25,750 & 0.45591 & 1.000 \\
        Claude Fable 5.1 & $\mathbf{76.5}\!\pm\!8.4$ & $73.6\!\pm\!9.4$ & $75.0\!\pm\!8.9$ & $67.9\!\pm\!11.7$ & $81.5\!\pm\!7.4$ & 10,400 & 8,200 & 5,500 & 24,100 & 0.30484 & 0.669 \\
        Claude Opus 5.5 & $75.1\!\pm\!8.7$ & $72.7\!\pm\!9.6$ & $73.9\!\pm\!9.1$ & $66.5\!\pm\!12.0$ & $81.0\!\pm\!7.6$ & 9,200 & 6,900 & 4,700 & 20,800 & 0.13218 & 0.290 \\
        GPT-6.1 Sol & $74.2\!\pm\!8.9$ & $72.0\!\pm\!9.7$ & $73.1\!\pm\!9.3$ & $65.1\!\pm\!12.2$ & $80.0\!\pm\!7.8$ & 9,200 & 6,000 & 4,400 & 19,600 & 0.06300 & 0.138 \\
        GPT-6.1 Terra & $72.9\!\pm\!9.1$ & $70.9\!\pm\!10.0$ & $71.9\!\pm\!9.5$ & $63.7\!\pm\!12.6$ & $78.5\!\pm\!8.0$ & 10,900 & 6,700 & 3,300 & 20,900 & 0.02774 & 0.061 \\
        GPT-5.6 Sol & $73.0\!\pm\!9.2$ & $70.5\!\pm\!10.0$ & $71.7\!\pm\!9.6$ & $63.4\!\pm\!12.6$ & $78.7\!\pm\!8.0$ & 11,400 & 7,000 & 5,200 & 23,600 & 0.15240 & 0.334 \\
        Claude Sonnet 5.5 & $73.2\!\pm\!9.1$ & $70.4\!\pm\!10.1$ & $71.7\!\pm\!9.6$ & $63.3\!\pm\!12.6$ & $79.0\!\pm\!8.0$ & 10,100 & 6,900 & 3,800 & 20,800 & 0.05889 & 0.129 \\
        GPT-5.6 Terra & $70.7\!\pm\!9.5$ & $68.9\!\pm\!10.4$ & $69.8\!\pm\!10.0$ & $61.0\!\pm\!13.1$ & $76.6\!\pm\!8.4$ & 13,000 & 8,400 & 4,600 & 26,000 & 0.08288 & 0.182 \\
        \bottomrule
    \end{tabular*}
\end{table*}

\begin{table*}[tp]
    \centering
    \footnotesize
    \setlength{\tabcolsep}{2pt}
    \caption{Component ablation with GPT-6 Astra across nine datasets. $\Delta F_1$ and $\Delta$ mIoU are gains over Agent only in percentage points, computed from the reported means. Bold marks the highest mean per metric.}
    \label{tab:ablation}
    \begin{tabular*}{\textwidth}{@{\extracolsep{\fill}}lccc*{7}{r}@{}}
        \toprule
        Configuration & S & I & V & $P\uparrow$ & $R\uparrow$ & $F_1\uparrow$ & mIoU $\uparrow$ & Type $\uparrow$ & $\Delta F_1$ & $\Delta$ mIoU \\
        \midrule
        Agent only & -- & -- & -- & $62.5\!\pm\!10.2$ & $59.4\!\pm\!11.1$ & $60.9\!\pm\!10.7$ & $45.1\!\pm\!14.4$ & $64.2\!\pm\!9.5$ & -- & -- \\
        Agent + S & $\checkmark$ & -- & -- & $66.0\!\pm\!9.8$ & $63.6\!\pm\!10.6$ & $64.8\!\pm\!10.2$ & $48.7\!\pm\!14.0$ & $68.0\!\pm\!9.1$ & +3.9 & +3.6 \\
        Agent + I & -- & $\checkmark$ & -- & $74.1\!\pm\!9.0$ & $73.0\!\pm\!9.4$ & $73.5\!\pm\!9.2$ & $63.6\!\pm\!12.3$ & $78.0\!\pm\!7.8$ & +12.6 & +18.5 \\
        Agent + I + V & -- & $\checkmark$ & $\checkmark$ & $\mathbf{79.1}\!\pm\!8.2$ & $\mathbf{77.9}\!\pm\!8.7$ & $\mathbf{78.5}\!\pm\!8.4$ & $\mathbf{73.2}\!\pm\!10.7$ & $\mathbf{85.0}\!\pm\!6.8$ & +17.6 & +28.1 \\
        Agent + S + I & $\checkmark$ & $\checkmark$ & -- & $70.7\!\pm\!9.2$ & $68.7\!\pm\!9.9$ & $69.7\!\pm\!9.6$ & $58.2\!\pm\!12.8$ & $74.7\!\pm\!8.3$ & +8.8 & +13.1 \\
        Full SERA & $\checkmark$ & $\checkmark$ & $\checkmark$ & $76.0\!\pm\!8.6$ & $74.2\!\pm\!9.3$ & $75.1\!\pm\!8.9$ & $68.4\!\pm\!11.6$ & $82.0\!\pm\!7.4$ & +14.2 & +23.3 \\
        \bottomrule
    \end{tabular*}
\end{table*}

\noindent\textbf{Backbone comparison.}
Table~\ref{tab:backbones} compares eight LLM backbones within SERA under fixed inputs, prompts, tools, and interaction budgets. GPT-6 Astra and Claude Fable 5.1 achieve comparable overall performance, with Astra attaining slightly better detection and localization results and Fable achieving marginally higher precision. More broadly, stronger backbones generally yield better performance, suggesting that SERA benefits from the capabilities of its underlying reviewer and verifier. This trend indicates room for further improvement as LLM agents become more capable: advances in reasoning and tool use could enable SERA to detect, localize, and characterize errors more effectively within the existing framework.

\noindent\textbf{Inference cost.}
Token counts and API costs are reported per reviewed slice, excluding slices filtered by local screening. Costs are calculated using uncached input, cache-read, and output token rates, excluding local processing costs. Costs are averaged before token counts are rounded.

Table~\ref{tab:backbones} highlights the trade-off between accuracy and API cost across backbones. GPT-6 Astra achieves the best overall performance, while Claude Fable 5.1 reduces cost by 33.1\% with only a 0.1-percentage-point drop in $F_1$. GPT-6.1 Sol and Terra offer further savings with moderate accuracy reductions. Notably, Sol uses the fewest tokens, whereas Terra incurs the lowest cost. These results demonstrate that SERA supports flexible backbone selection based on accuracy and budget requirements.

\subsection{Ablation Study}
\label{sec:ablation}

\noindent\textbf{Configurations.}
Table~\ref{tab:ablation} compares six configurations using GPT-6 Astra under the same evaluation protocol. The three components are feature-based slice screening (S), box initialization with organ profiles (I), and agent verification and box refinement (V). Configurations without S review every slice with the agent, whereas configurations with S review only the selected slices. These settings allow matched comparisons of verification with and without screening, as well as screening with different combinations of the other components. Full SERA combines all three components.

\noindent\textbf{Initialization and verification.}
The results highlight the complementary roles of initialization and verification with refinement. Initialization substantially improves detection and localization, with a particularly pronounced gain in box mIoU, suggesting that an effective starting point is important for accurately identifying error regions. Verification and refinement provide further improvements, indicating that subsequent checking and correction remain valuable even after initialization. Similar gains with and without screening suggest this contribution is consistent across both settings. Agent + I + V achieves the highest mean scores across all five metrics, while the improvements to Agent + S + I show that verification and refinement also strengthen the complete SERA pipeline.

\noindent\textbf{Effect of slice screening.}
Screening introduces an efficiency–accuracy tradeoff, and its balance depends on the agent configuration. Its modest benefit to Agent-only does not persist once initialization is available, suggesting that early filtering can constrain the gains from a more capable review pipeline. The reduction in recall further highlights the potential cost of excluding slices before agent review. Although removing screening achieves the highest predictive quality, reviewing every slice consumes approximately 30 times as many tokens per volume as Full SERA. Including screening therefore offers a practical compromise: Full SERA retains most of the detection and localization gains from initialization and verification while substantially reducing token consumption. These results support screening as a way to make agent review more affordable at scale.

\section{Conclusion}
\label{sec:conclusion}

Localized errors often remain unnoticed in otherwise plausible segmentations, while manual review is time-consuming. To address this challenge, we propose SERA, an evidence-guided LLM-agent framework that detects and localizes segmentation errors. SERA combines organ-specific initialization with agent-driven visual inspection and tissue measurements to identify and refine error regions. Efficient screening further reduces token costs. By providing error locations, types, and supporting evidence, SERA provides interpretable, evidence-based insights into segmentation errors.

\section*{LLM Usage Statement}
LLMs were used as agent backbones in the proposed SERA framework for segmentation error analysis and verification. AI-assisted writing tools were used solely for grammar checking and language refinement. All scientific ideas, methodology, experiments, analyses, and manuscript content were developed, reviewed, and verified by the authors.

\FloatBarrier
\clearpage
\setcounter{page}{1}
\appendix
\setcounter{figure}{0}
\renewcommand{\thefigure}{S\arabic{figure}}
\renewcommand{\theHfigure}{supp.\arabic{figure}}
\twocolumn[{%
  \centering
  \pdfbookmark[0]{Supplementary Material}{supplement}
  \Large\textbf{\papertitle}\\
  \vspace{0.5em}Supplementary Material\\
  \vspace{1em}}]

\begin{strip}
\centering
\includegraphics[width=\textwidth]{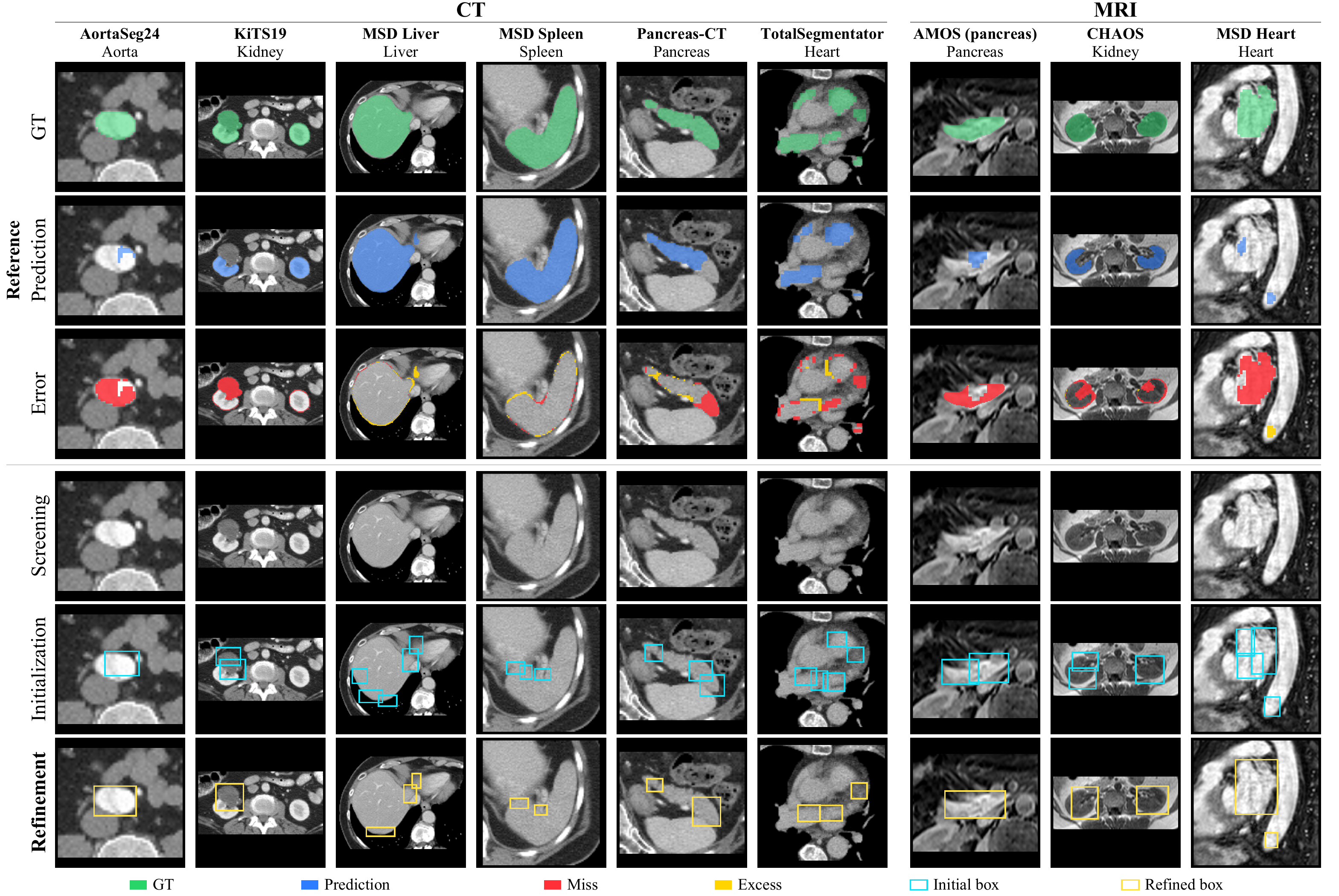}
\captionsetup{hypcap=false}
\captionof{figure}{\textbf{Illustration of segmentation error review across nine datasets.}
The top rows show the reference mask (green), input mask (blue), and their differences (red for missing tissue and yellow for excess tissue), while the bottom rows illustrate the results at different stages of SERA.
}
\label{fig:supp_stage_outputs}
\captionsetup{hypcap=true}
\end{strip}

\section{Outputs at each stage}
\label{sec:supp_stage_outputs}

\noindent\textbf{Slice screening.}
SERA first returns the indices of axial slices that warrant inspection. It ranks slices by local signs of mask error and changes across neighboring slices, then suppresses redundant views of the same candidate.

\smallskip
\noindent\textbf{Box initialization.}
SERA groups signs of missing or excess tissue into regions and fits an initial box to each using an organ-specific profile. Small irregularities along a mask boundary can generate several nearby proposals. Some capture minor contour differences, while others cover different parts of the same error.

\newpage
\noindent\textbf{Verification and refinement.}
The agent inspects image crops, tissue measurements, and neighboring slices to decide which candidates warrant a finding and how far each box should extend. It can set aside a minor contour discrepancy with little support on adjacent slices and retain a clearer omission that persists across them. A verifier reexamines uncertain findings and can restore rejected candidates. Consolidating overlapping findings then focuses the report on distinct errors. In the KiTS19 illustration, one final box spans the region covered by the two proposals. Each finding gives its slice, box coordinates, error direction, and supporting evidence.

\clearpage
\section{What the agent reads and returns}
\label{sec:supp_agent_evidence}

\noindent\textbf{Input records.}
Alongside the numbered slice overview, the reviewing agent receives a JSON object with the slice index (\texttt{slice\_z}), image dimensions, voxel spacing, and input-mask area. Its \texttt{candidates} list contains one record per numbered region. Each record gives the candidate ID, box coordinates $(x_0,y_0,x_1,y_1)$ on the full slice, whether the feature lies inside or outside the mask, and the area supporting each feature. A nested \texttt{features} record describes the candidate's tissue intensity, its contact with the mask, and the image gradient at that interface. On CT, intensities are in HU and gradients in HU/mm. MRI uses normalized intensities and their gradients per millimeter.

\smallskip
\noindent\textbf{Interpreting measurements.}
The agent compares the candidate's median intensity and percentile range with those of the mask core. It also compares the median gradient where the candidate touches the mask with the median along the mask boundary on the same slice. These comparisons help assess whether the contour follows a visible tissue interface or passes through tissue with similar intensity on both sides. Neighboring slices and orthogonal views further check whether the tissue connects to the target or to an adjacent structure.

\smallskip
\noindent\textbf{Tool requests.}
Each review turn produces a JSON object with a short \texttt{observations} field and a list of \texttt{tool\_calls}. Each request specifies the tool, the candidate ID or image region, and any parameters needed for the operation. The agent can inspect additional views, sample an intensity profile across the contour, compare the candidate with neighboring tissues, or adjust the box extent. The tool returns images, measurements, or updated coordinates linked to the candidate for use in the next turn.

\smallskip
\noindent\textbf{Final report.}
On the final turn, \texttt{tool\_calls} is empty and \texttt{final} lists findings and any explicit rejections. Each finding specifies the candidate ID, error direction, confidence, image and measurement evidence, strongest alternative explanation, and a review instruction. The agent attaches the final box coordinates. For each rejected candidate, the agent records the evidence supporting its decision.

\end{document}